# Automated Noncontact Trapping of Moving Micro-particle with Ultrasonic Phased Array System and Microscopic Vision

Mingyue Wang, Jiaqi Li, Yuyu Jia, Zhenhuan Sun, Yuhang Liu, Teng Li*, and Song Liu*, *Member, IEEE*

*Abstract*—Noncontact particle manipulation (NPM) technology has significantly extended mankind's analysis capability into micro and nano scale, which in turn greatly promoted the development of material science and life science. Though NPM by means of electric, magnetic, and optical field has achieved great success, from the robotic perspective, it is still labor-intensive manipulation since professional human assistance is somehow mandatory in early preparation stage. Therefore, developing automated noncontact trapping of moving particles is worthwhile, particularly for applications where particle samples are rare, fragile or contact sensitive. Taking advantage of latest dynamic acoustic field modulating technology, and particularly by virtue of the great scalability of acoustic manipulation from micro-scale to sub-centimeter-scale, we propose an automated noncontact trapping of moving micro-particles with ultrasonic phased array system and microscopic vision in this paper. The main contribution of this work is for the first time, as far as we know, we achieved fully automated moving micro-particle trapping in acoustic NPM field by resorting to robotic approach. In short, the particle moving status is observed and predicted by binocular microscopic vision system, by referring to which the acoustic trapping zone is calculated and generated to capture and stably hold the particle. The problem of hand-eye relationship of noncontact robotic end-effector is also solved in this work. Experiments demonstrated the effectiveness of this work.

## I. INTRODUCTION

Noncontact particle manipulation (NPM) can effectively prevent heterogeneous nucleation and contamination from a container wall, so it has received a lot of interest in fundamental science fields such as physics [1], chemistry [2], biology [3], crystallography [4], medical science [5], etc. Since the invention of optical tweezer by Arthur Ashkin who realized micro-particle manipulation through a focused laser beam [6], scientists and researchers went deeper into this technology and achieved great progress of noncontact micro-particle manipulation in different mediums such as liquid and gas through various physical fields such as optical field [7], plasma field [8], and magnetic field [9]. For example, magnetic tweezers can easily stretch and rotate magnetic spheres through the change of external magnetic field, so they can be applied in the fields related to DNA molecular manipulation [10]. As a versatile manipulation platform, acoustic manipulation has attracted great attention since *Wu* and *Du* initially trapped frog egg clusters through a potential well generated by two reverse collimated focused ultrasonic beams in water [11]. Benefitting from the unique characteristic of excellent bio-compatibility, high resolution, and strong penetrability, the latest acoustic manipulation has already achieved in-vivo micro particle manipulation in living body using ultrasound beams of specific shapes produced in a phased array [12].

Acoustic manipulation is primarily realized by resorting to acoustic streaming [13-14] or acoustic radiation force [15], since acoustic waves transfer angular momentum and linear momentum along propagation. Acoustic streaming has been successfully applied to particle manipulation like cell incubation, droplet microfluidics, protein engineering, etc. For example, *Wixforth et al.* designed a programmable nano-pump on a piezoelectric chip for small droplets manipulation by surface acoustic wave (SAW) [16]. *Schmid et al.* reported a versatile microfluidic fluorescence-activated cell sorter using travelling SAW induced streaming [17]. Acoustic radiation force enables precision control over particle position. By the acoustic impedance contrast between particle and propagation medium, acoustic radiation force shows different properties [18-19]. In [20-21], *Yang et al.* used an ultrasonic transducer array system to achieve 3-D levitation and translation of polydimethylsiloxane (PDMS) particles and polystyrene (PS) particles along pre-loaded user-customized motion trajectory.

We herein categorize acoustic manipulation, by working mechanism, as static [22] versus dynamic [23], depending on whether the acoustic field can be modulated according to different working scenarios or manipulation purposes. From the robotic perspective, automated acoustic manipulation, which inherently is dynamic acoustic manipulation, would greatly promote the robotic manipulation capability into non-contact, non-invasive and in-vivo applications. As dynamic acoustic modulation technologies like ultrasonic phased array emerged [24] in recent years, acoustic robotic manipulation is now worthy our efforts.

In this work, we report our latest research towards robotic acoustic manipulation regarding automated noncontact trapping of moving micro-particle with ultrasonic phased array system and microscopic vision. The fully automated system, through our numerical calculation and physical experiment validation, is capable of trapping and firmly hold both PDMS and PS particles. This work is expected to greatly benefit NPM applications where particle samples are rare, fragile or contact sensitive. Details regarding the system setup, hand-eye relationship calibration, trapping strategy, and experiment results are given in rest of the paper.

This paper was sponsored by Shanghai Pujiang Program under Grant 21PJ1410500. (*Mingyue Wang and Jiaqi Li Contribute equally to this work*) (*Corresponding Author: Teng Li and Song Liu*)

M. Wang, J. Li, Y. Jia, Z. Sun, and T. Li are with the School of Information Science and Technology, ShanghaiTech University, Shanghai 201210, China (e-mail: wangmy1, lijq1, sunzhh, jiayy1, liteng1@shanghaitech.edu.cn).

Y. Liu is with School of Computer Science and Technology, Shandong Jianzhu University, Jinan 250200, China (olivia_liuyuhang@outlook.com).

S. Liu is with the School of Information Science and Technology, ShanghaiTech University, Shanghai 201210, China, and with Shanghai Engineering Research Center of Intelligent Vision and Imaging, Shanghai, China (e-mail: liusong@shanghaitech.edu.cn ).

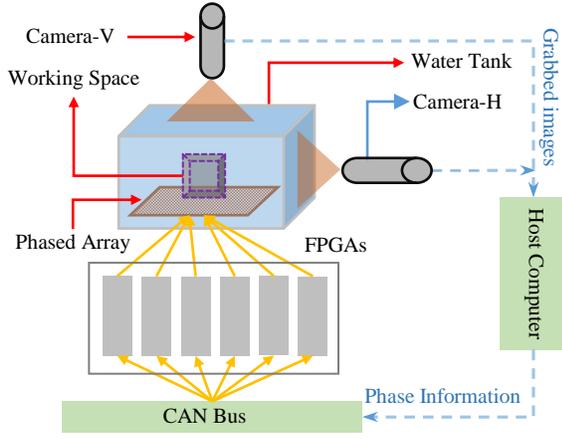

Fig. 1. Schematic overview of the automated acoustic manipulation platform. The system consists of an ultrasonic phased array system as its end-effector and binocular microscopic vision system for particle posture feedback. The acoustic field is modulated by controlling the sinusoidal driving voltage phases implemented through FPGA.

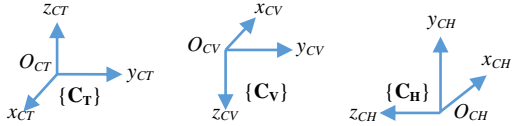

Fig. 2 Coordinates establishment of the automated acoustic manipulation platform, i.e., one phased array coordinates and two vision coordinates.

## II. PLATFORM CONFIGURATION AND CALIBRATION

### A. Acoustic NPM Platform Configuration

The automated acoustic NPM platform is configured as Fig. 1, which primarily consists of an ultrasonic phased array system, binocular microscopic vision system, and a host computer. In specific, a DI water tank sits on top of the ultrasonic phased array. Two optical microscopic cameras are mounted orthogonally as top and side view to the water tank, whose field of view (FOV) intersection defines the working space. The host computer takes charge of controlling the entire trapping process, including capturing live images from cameras, processing image, detecting and predicting particle position, calculating and sending phase data to phased array. Customized FPGAs receive the phase data and generate phase-modulated sinusoidal driving voltages to phased array in order to generate an acoustic trapping zone on particle position.

The coordinates are established as Fig. 2. Coordinates $\{C_T\}$ is established top left corner on the ultrasonic phased array. Coordinates $\{C_V\}$ and $\{C_H\}$ are established on the imaging planes of corresponding microscopic cameras with origins being the intersection of optical axes and imaging planes. The $x$- and $y$-axes of $\{C_V\}$ and $\{C_H\}$ follow the direction of $u$- and $v$-axes of their images while $z$-axes follow the direction of optical axes.

### B. Eye-to-Hand Relationship Calibration

The main challenge of non-contact robotic manipulation, compared with conventional counterpart, is the end-effector, which in this case is actually the acoustic field, is invisible. In this section, we articulate a solution to this invisible eye-to-hand relationship calibration problem by referring to Image Jacobian matrix and reference points [25].

Image Jacobian matrix is the transformation matrix between motion increments from Cartesian coordinates to image coordinates. In this work, we have

$$[\Delta u_H, \; \Delta v_H, \; \Delta u_V, \; \Delta v_V]^T = J[\Delta x, \; \Delta y, \; \Delta z]^T \quad (1)$$

where $[\Delta x, \Delta y, \Delta z]^T$ represents particle motion increment in $\{C_T\}$, and $[\Delta u_H, \Delta v_H, \Delta u_V, \Delta v_V]^T$ represents feature motion increment in $\{C_H\}$ and $\{C_V\}$. Calibration method of $J$ is given in [26] in detail.

Image Jacobian matrix is essentially regarding vector transformation from motion space to vision space. However, to achieve automated noncontact trapping through invisible robot end-effector, absolute particle position in $\{C_T\}$ is mandatory for localization of trapping zone. By referring to reference points in working space, whose coordinates in $\{C_T\}$, $\{C_H\}$ and $\{C_V\}$ are all known in advance, the particle can be localized from microscopes in $\{C_T\}$ as

$$\begin{bmatrix} x_{ptc} \\ y_{ptc} \\ z_{ptc} \end{bmatrix} = J^+ \left( \begin{bmatrix} u_{H-ptc} \\ v_{H-tpc} \\ u_{V-ptc} \\ v_{V-ptc} \end{bmatrix} - \frac{1}{n}\sum \begin{bmatrix} u_{H-ref} \\ v_{H-ref} \\ u_{V-ref} \\ v_{V-ref} \end{bmatrix} \right) + \frac{1}{n}\sum \begin{bmatrix} x_{ref} \\ y_{ref} \\ z_{ref} \end{bmatrix} \quad (2)$$

where subscript of *ref* means coordinates of reference points, while *ptc* means coordinates of particle under manipulation, *n* is number of reference points. In practice, reference points are obtained by first generating acoustic field with a single focal point somewhere in working space like Fig. 3, then scanning the acoustic field with a needle hydrophone to localize the hydrophone at maximum pressure, and finally extracting image coordinates of hydrophone needle tip.

## III. ACOUSTIC TRAP ZONE DESIGN AND GENERATION

### A. Acoustic Trap Design

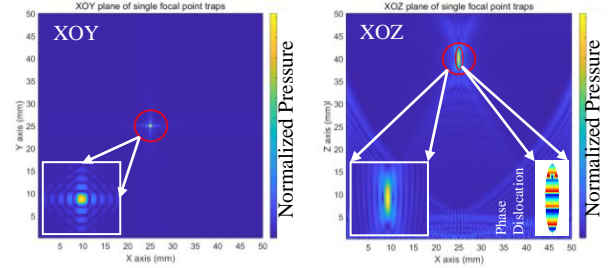

Fig. 3. Simulation results by COMSOL Multiphysics regarding acoustic pressure distribution of a single focal point trap at $[25, 25, 40]^T$ on both XOY plane and XOZ plane, which forms a trapping zone for particles with negative acoustic impedance contrast, like PDMS particle.

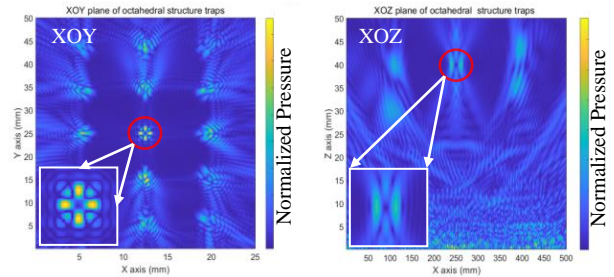

Fig. 4. Simulation results by COMSOL Multiphysics regarding acoustic pressure distribution of an **octahedral trap** at $[25, 25, 40]^T$ on both XOY plane and XOZ plane, which forms a trapping zone for a particle with positive acoustic impedance contrast, like PS particle.

The acoustic impedance contrast of particle material against propagation medium results in different properties of the acoustic radiation force, by which particles are categorized as negative impedance contrast and positive impedance contrast. A simple single acoustic focal point [22] is well enough to generate a trapping zone, as shown by Fig. 3, for particles with negative impedance contrast, like PDMS particle. For particles with positive impedance contrast, like PS particle, a trapping zone centered with low pressure while surrounded with high pressure, like a vortex trap, must be generated [21]. In this paper, we propose a novel octahedral trap, as shown by Fig. 4, supporting real-time phase data calculation with firm trapping capability.

The proposed octahedral trap is designed and characterized by six vertexes evenly sampled from a 3D sphere. Once the six vertexes are assigned high acoustic pressure, a central null pressure area fundamentally functioning as trapping zone for particles with positive impedance contrast, will be formed in the geometrical center of the vertexes. The sphere diameter is the tuning parameter of the octahedral trap, which should be optimized per the phased array design to obtain the best pressure contrast. The single focal point trap and the octahedral trap are capable of stably trapping particles of different acoustic properties, and provide enough acoustic radiation force in all three directions. Further automated and precision acoustic manipulation with the traps, like translating and rotating, are investigated in future efforts.

*B. Acoustic Traps Generation*

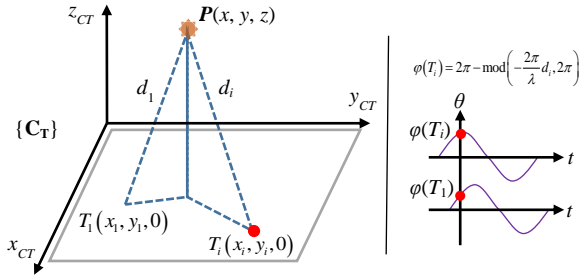

Fig. 5. Schematic depiction of generating single focal point trap for particles with negative acoustic impedance contrast. Transducers in phased array are modeled as piston source labeled by their geometrical center coordinates $T_i$, whose initial oscillation phases are modulated by aligning the phase delay towards focal position to a specific value, like $2\pi$.

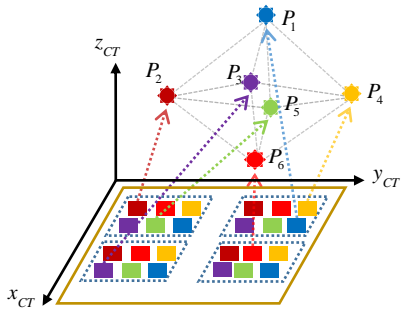

Fig. 6. Schematic depiction of generating octahedral trap for particles with positive acoustic impedance contrast. The octahedral trap is formed by generating focal zones at six vertexes through spatial multiplexing (**SM**) method. By taking advantage of high transducer density of high space-bandwidth product of phased array, transducers are grouped to take charge of generating different focal points. The color map depicts grouping method.

The ultrasonic phased array system is capable of controlling the acoustic field through phase modulation of sinusoidal driving voltage to the transducers. The phase data essentially composes a phase-only hologram (POH). State-of-the-art method for POH calculation as per an expected acoustic field in 3D space is iterative backpropagation (IB) algorithm [12], which can converge to a local optimum by chance through iteration. However, as an automated robotic trapping system, real time response of a closed-loop system is of great importance. Thus, methods based on iteration approach are not suitable for real-time tasks. Here in this work, by referring to Fresnel half-wave bands [22] theory, the phase $\varphi(T_i)$ of a transducer $T_i$ contributing positively to the pressure of a focal point $P$ is calculated by

$$\varphi(T_i) = 2\pi - \mathrm{mod}\left(-\frac{2\pi}{\lambda}d_i, 2\pi\right) \qquad (3)$$

where $d_i$ is the Euclidean distance between a focal point $P$ and transducer $T_i$, and $\lambda$ is wave length of sound in water under room temperature. Details are given Fig. 5.

For single focal point trap at point $P$, directly applying Eq. (3) to each transducer element in the phased array yields the POH. As regarding the novel octahedral trap described in last section by Fig. 4, we propose a novel and effective spatial-multiplexing (SM) method, which supports real-time calculation. In SM, as in Fig. 6, the transducers in phased array are grouped into blocks of $2 \times 3$ in size, in which each element contributes to a pre-assigned focal point. The SM method may lose validity if the phased array has low space-bandwidth product (SBP) [18] or low resolution, but is validated effective in this work ($50 \times 50$ phased array, 1 mm transducer size, working at 2.3 MHz).

## IV. AUTOMATED TRAPPING STRATEGY

*A. Feature Extraction*

In this work, both PS and PDMS particles are spheres with the concerned image feature being the geometrical center. Image processing workflow is given in Fig. 7. Generally, with grabbed images from stereo microscopes, background subtraction and adaptive binarization is first applied so that the noises in background are filtered while foreground information persisted. Then, roughly localize the particle in the image through counting the number of foreground pixels in a sliding window depending on prior knowledge of particle size. Afterwards, apply erosion-after-dilation morphological processing to the sliding window for further filtering against illumination. Finally, the contour can be retrieved from the binary image and fitted to ellipse by RANSAC. This image processing workflow is dedicatedly articulated since its real-time performance is critical for successful automated trapping of moving particles.

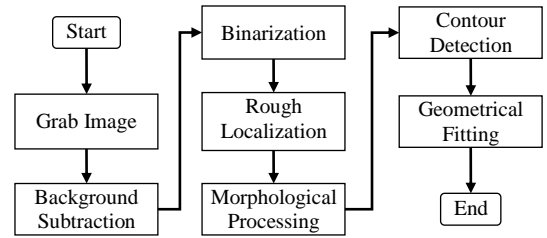

Fig. 7. Real-time feature extraction flowchart.

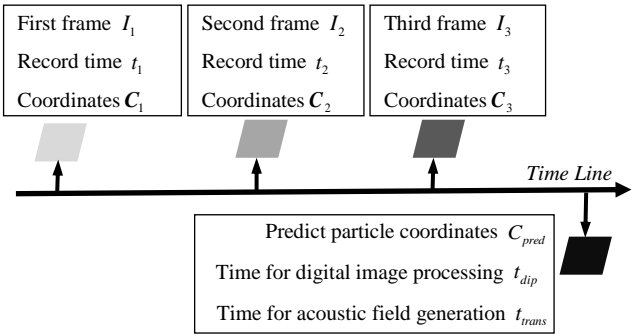

Fig. 8. Illustration of moving particle position prediction along timeline.

*B. Moving Particle Position Prediction*

The particle motion status prediction is necessary since digital image processing and acoustic field generating yields time delay. In other words, an acoustic trap must be generated exactly at the moment at one specific position where one particle shows up. Here in this work, without losing generality, the particles under manipulation are considered to follow uniform linear motion. Therefore, three successive frames (each frame contains two images from camera-H and camera-V simultaneously) are analyzed, as in Fig. 8, to predict particle position at a specific time by

$$C_{pred} = C_3 + \frac{C_3 - C_1}{t_3 - t_1} \times (t_{dip} + t_{trans}) \quad (4)$$

where particle coordinates $C_1$, $C_2$ and $C_3$ are obtained through pseudo-inverse of image Jacobian matrix with the detected image features from images, $t_{dip}$ is the time for digital image processing which can be get directly from the program, and $t_{trans}$ is the time for phase calculation, phase data transmission, and acoustic field generation, which is estimated in advance and set manually.

*C. Automated Trapping Strategy*

We herein argue that the automated trapping of moving particle showing sophisticated motion trajectories (either in position or velocity) is not the focus of this work, though it may be critical in some particular application scenarios. Instead, we are proposing our efforts towards real-time robotic acoustic manipulation aiming to push forward this cutting-edge NPM technology.

The control diagram for automated acoustic trapping of moving particles is shown in Fig. 9. Since the particle material acoustic properties can't be detected automatically, positive or negative impedance contrast material should be ***first*** manually selected before automated trapping starts up. ***Then*** images are successively grabbed and analyzed until particle position can be well predicted given all the previous information including position and time by Eq. (4). ***Afterwards***, with the predicted particle position, phase data to drive phased array are calculated by the scheme discussed in Section III, and ***finally*** sent from host computer to FPGAs to generate corresponding driving sinusoidal voltage in order to form an acoustic trapping zone at target position. The trapping task will not be terminated until the particle position is well confirmed to be within the trapping zone. Since trapping zone is larger than particle size, automatic acoustic trapping shows robustness to localization errors.

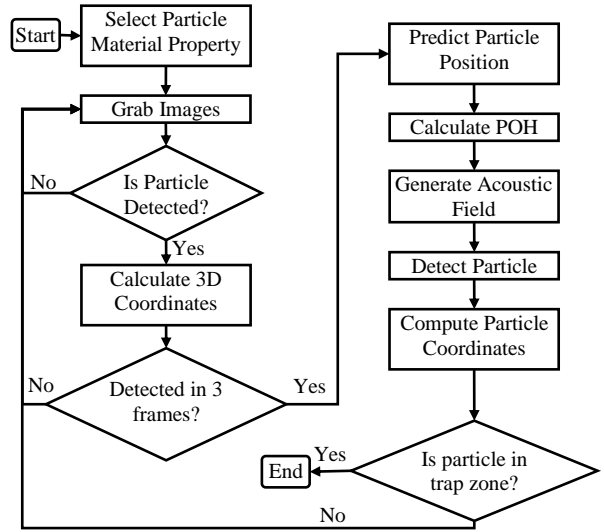

Fig. 9. The control diagram for automated acoustic trapping of moving particles with either positive or negative acoustic impedance contrast.

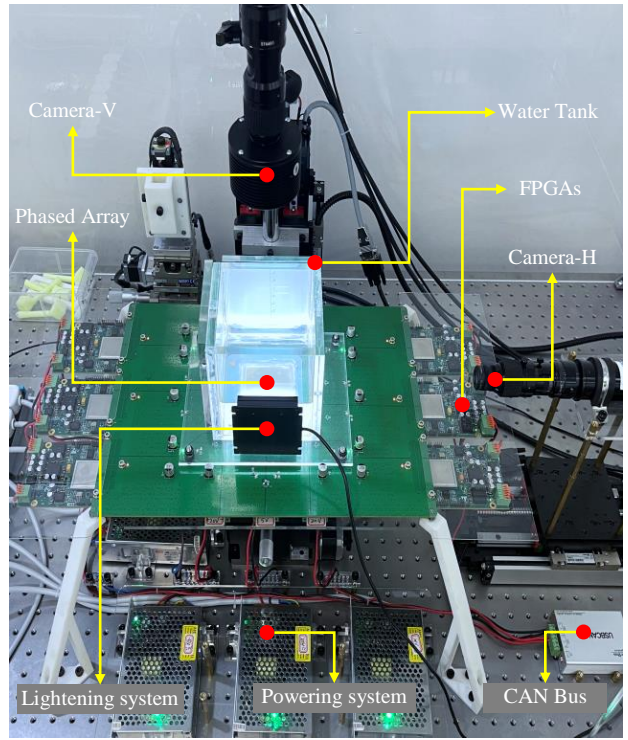

Fig. 10. Experimental system setup for robotic acoustic manipulation.

## V. EXPERIMENTS AND RESULTS

*A. Experiment System Setup*

An experiment system was established according to the system configuration given in Section II-A, as shown in Fig. 10. In this system, the ultrasonic phased array comprises 2500 elements arranged as $50 \times 50$ square matrix structure working at frequency of 2.3 MHz with peak-to-peak driving voltage of 5 V. The phased array was fabricated on a bulk lead zirconate titanate (PZT) ceramic with size of $50 \times 50$ mm$^2$ using MEMS technology. The water tank is made from transparent acrylic board sitting on top of PZT storing deionized water. The binocular vision system is composed of two GC2450 microscopic cameras, equipped with Navitar zoom lens, capable of capturing images 15 frames per second (FPS) by size of 2448 ×2050 in pixel. The two orthogonally mounted cameras share a common FOV about $37 \times 30 \times 30$ mm$^3$ in size right above the phased array, which defines the working space. The CAN-Bus controller works at baud rate of 500 kbps, enabling POH updating at 11 PFS.

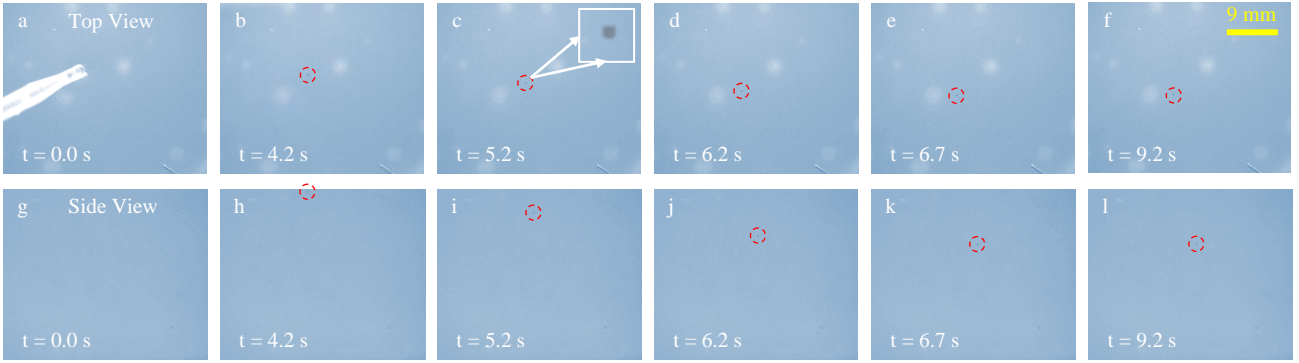

Fig. 12. Sequential images captured from microscopes during the trapping process of a PS particle in one experiment trial. The particle was released from dropping pipette to the water tank at start. At the moments of 4.2 s, 5.2 s and 6.2s, the particle was freely falling, and was firmly trapped since 6.7 s. Images at 9.2 s were to validate the particle was firmly trapped. Particle size was 400 μm. Images were enhanced in contrast for better clarity.

### B. Eye-to-Hand Relationship Calibration

Eye-to-Hand relationship calibration concerns with localization of feature points in both manipulation space and vision space, from which we get the motion increments to calibrate image Jacobian matrix J, and determine the reference points. As mentioned in Section II.B, the feature points are visualized by referring to a needle hydrophone (Precision Acoustics Ltd., UK, tip size 200 μm). Therefore, successful calibration well depends on feature point localization accuracy. Fig. 11 shows the cross-sectional pressure profile of a single focal point trap obtained by scanning the acoustic field with the hydrophone, from which we can see that the highest pressure shows up at the focal center. However, considering the 200 μm hydrophone need tip size and taking the image processing errors in to account, there inevitably are calibration errors consequentially resulting in particle localization errors in practice.

$$J = \begin{pmatrix} -0.0002 & -0.0631 & -0.0009 \\ -0.0001 & 0.0012 & -0.0634 \\ -0.0623 & -0.0044 & 0.0003 \\ 0.0043 & -0.0623 & 0.0011 \end{pmatrix} \; pixel/\mu m, \quad (5)$$

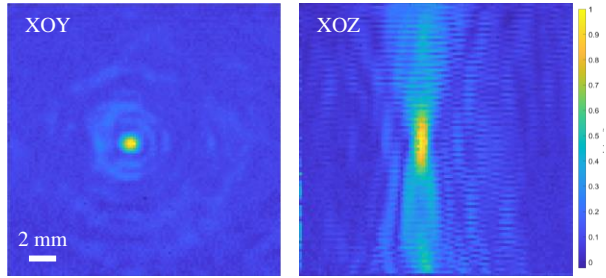

Fig. 11. The cross-sectional pressure profile of a single focal point trap obtained from physically scanning the acoustic field with a needle hydrophone. Both hydrophone needle tip and scanning size was 200 μm.

The finally calibrated image Jacobian matrix J was based on 24 reference points, whose centroid was $[25, 25, 40]^T$ mm in $\{C_T\}$, $[854.2, 951.4]^T$ pixel in $\{C_V\}$ and $[1328.1, 716.4]^T$ pixel in $\{C_H\}$ respectively. With the calibrated eye-to-hand relationship, the particle localization errors were evaluated through experiments to be less than 100 μm characterized by Euclid distance, which was about one seventh to a quarter of the particle diameter used in this work, or one third of the half-wavelength (i.e., diffraction limit). This localization accuracy was both theoretically and practically acceptable.

### C. Automated Trapping of Moving Particles

The automated noncontact trapping experiments of freely falling PS and PDMS particles with size from 400 μm to 700 μm in diameter were well conducted. Complicated scenarios like where propagation medium showing up turmoil or the particle exhibiting great mobility are out of the scope of this paper. Through extensive experiments, both PS and PDMS particles can be captured and firmly held from moving status, with a successful trapping rate of more than 80%. We herein argue that the successful trapping rate can be further improved by more advanced digital image processing technique, more sophisticated particle motion prediction method and optimization of the trapping strategy. Details regarding the experiment results are given as follows.

Fig. 12 shows the sequential images captured from the microscopes during the trapping process of a 400 μm PS particle in one experiment. The particle was released from dropping pipette to the water tank and started freely falling. Then the system was triggered to detect the particle from microscopes. Once the particle was detected simultaneously from microscopes for three successive times, particle position at 150 ms (including image processing time for 60 ms and phase data sending time from host computer to FPGAs for 90 ms) in advanced was predicted. Thereafter, an octahedral trap was designed by which the POH to phased array was calculated. Finally, the phase data was delivered and implemented to generate a trapping zone at the expected position. The particle was firmly trapped since 6.7 s. In this experiment, it took about 2.5 s to trap the PS particle. This is for the reason that for better reader's clarity, the system was customized to trap the particle somewhere around the FOV center of both microscopes.

TABLE I. RESULTS OF REPETITIVE AUTOMATED TRAPPING EXPERIMENTS OVER PARTICLE WITH DIFFERENT MATERIALS AND SIZES

| Exp. No. | Particle type | | Stable Trapping Time (s) | Position (mm, in $\{C_T\}$) | | |
|---|---|---|---|---|---|---|
| | Material | Size (diameter) | | Particle Trapping position | Trapping Zone Position | Deviation |
| 1 | PDMS | 700 μm or 45 pixels | >60 | 28.0576, 17.6101, 42.6509 | 28.2429, 17.7410, 41.0819 | 1.585317602 |
| 2 | | | >60 | 26.6736, 10.4052, 40.7055 | 26.8998, 10.8456, 38.7462 | 2.02088473 |
| 3 | | | 15 | 22.6789, 20.3556, 46.6741 | 22.9254, 20.8502, 44.2527 | 2.483660478 |
| 4 | | 500 μm or 31 pixels | 50 | 23.0663, 19.3140, 43.0424 | 23.1535, 19.5085, 40.7328 | 2.319415066 |
| 5 | | | >60 | 17.5401, 12.4632, 33.8870 | 18.3582, 12.5333, 31.9331 | 2.119416625 |
| 6 | PS | 520 μm or 33 pixels | >60 | 22.4580, 25.9707, 42.5061 | 22.6883, 25.9197, 42.5055 | 0.23588016 |
| 7 | | | >60 | 25.7273, 19.9716, 35.8733 | 25.8778, 19.9093, 35.8725 | 0.162887016 |
| 8 | | 400 μm or 25 pixels | >60 | 24.3404, 22.4867, 45.1019 | 24.5035, 22.3961, 45.1431 | 0.191069124 |
| 9 | | | >60 | 24.9879, 20.3091, 37.5523 | 25.1931, 20.3557, 37.5370 | 0.210980307 |
| 10 | | | >60 | 22.2190, 21.1474, 44.4188 | 22.4159, 21.1668, 44.5048 | 0.21573588 |

Repetitive experiments over different particle sizes were also conducted to validate and analyze the robustness of the proposed automated noncontact particle trapping approach. In all experiments, the octahedron from which we got the six vertexes to design the octahedral trap was 2.4 mm in diameter regardless of the particle size under manipulation. As can be seen from Table I, PS particles with diameter of 520 μm and 400 μm can both be firmly automated trapped. There existed deviation between the particle trapping position and the trapping zone position when particles were stably trapped. Through repetitive experiments, the deviation reached as high as 235 μm, which was acceptable from the reason that the diffraction limit is about 330 μm. Meanwhile, the water tank was $110 \times 110 \times 60$ mm$^3$ in size. Acoustic reflection due to propagation medium discontinuity in limited space would also cause disturbance to the particle position. This deviation can further be improved by customize the octahedral trap design per the size of PS particles.

Experiment results over PDMS particles with size of 700 μm and 500 μm were also articulated from Table I. It can be seen that both the trapping stability and the trapping position deviation is not comparable to that of PS particle case. This can be well explained by the periodic phase dislocation within the elliptic focal zone, as shown in Fig. 3, which substantially results in acoustic streaming upwards. A particle with negative impedance contrast, like the PDMS particle here in this case, will be stabilized where the acoustic radiation force balances the acoustic streaming induced drag force. This essentially proves the fact that a single focal point trap calculated from the simple Fresnel half-wave bands theory for single-sided phased array is not well suitable for particle trapping with negative impedance contrast. This also leads to one of our future endeavors to design a better acoustic trap by either eliminating the phase dislocation within the focal zone or reshaping the elliptic focal zone to be more compact, by either case real-time performance should be well supported.

In summary, we experimentally demonstrated that the automated noncontact acoustic trapping of moving particle is feasible by resorting to robotic approach, which in turn, as we expected, to push forward acoustic NPM technology.

## VI. CONCLUSION

The main contribution of this work is, for the first time, we achieved fully automated moving micro-particle trapping in acoustic NPM field based on our latest dynamic acoustic field modulating technology. In this work, we proposed a complete pipeline to fulfill automated acoustic trapping, including particle detection, particle motion prediction, phase calculation, and acoustic field generation. The image Jacobian matrix is used to solve the eye-to-hand relationship of invisible robot end-effector. In addition, benefited from the dynamic performance of our high-density ultrasonic phase array, the spatial-multiplexing (SM) method is applied to generate target acoustic field. We also proposed a novel octahedral trap for particle manipulation with positive impedance contrast. A series of experiment including trapping particles of different size and material were well conducted. The automated trapping of both PDMS particle and PS particle was achieved. The experimental results show the feasibility of applying robotic methods with acoustic NPM, which could potentially extend robotic manipulation capacity in micro- and nano scale.

## REFERENCES

[1] R. R. Dagastine, R. Manica, S. L. Carnie, D. Y. C. Chan, G. W. Stevens, and F. Grieser, "Dynamic forces between two deformable oil droplets in water", *Science*, vol. 313, no. 5784, pp. 210-213, 2006.

[2] S. Santesson, and S. Nilsson, "Airborne chemistry: acoustic levitation in chemical analysis", *Analytical and Bioanalytical Chemistry*, vol. 378, no. 7, pp. 1704-1709, 2004.

[3] M. D. Wang, H. Yin, R. Landick, J. Gelles, and S. M. Block, "Stretching DNA with optical tweezers", *Biophysical Journal*, vol. 72, no. 3, pp. 1335-1346, 1997.

[4] S. Tsujino, and T. Tomizaki, "Ultrasonic acoustic levitation for fast frame rate X-ray protein crystallography at room temperature", *Scientific reports*, vol. 6, no. 1, pp. 1-9, 2016.

[5] S. H. Kim, and K. Ishiyama, "Magnetic robot and manipulation for active-locomotion with targeted drug release", *IEEE Transactions on Mechatronics*, vol. 19, no. 5, pp. 1651-1659, 2014.

[6] A. Ashkin, J. M. Dziedzic, J. E. Bjorkholm, and S. Chu, "Observation of a single-beam gradient force optical trap for dielectric particles", *Optics letters*, vol.11, no.5, pp. 288-290, 1986.

[7] J. E. Curtis, B. A. Koss, and D. G. Grier, "Dynamic holographic optical tweezers", *Optics Communications*, vol. 207, no. 1-6, pp. 169-175, 2002.

[8] M. L. Juan, M. Righini, and R. Quidant, "Plasmon nano-optical tweezers", *Nature photonics*, vol. 5, no. 6, pp. 349–356, 2011.

[9] K. E. McCloskey, J. J. Chalmers, and M. Zborowski, "Magnetic cell separation: Characterization of magnetophoretic mobility", *Analytical chemistry*, vol. 75, no. 24, pp. 6868–6874, 2003.

[10] I. D. Vlaminck, and C. Dekker, "Recent advances in magnetic tweezers", *Annual Review of Biophysics*, vol. 41, no. 1, pp. 453-472, 2012.

[11] J. Wu, and G. Du, "Acoustic radiation force on a small compressible sphere in a focused beam", *The Journal of the Acoustical Society of America*, vol. 87, no. 3, pp. 997–1003, 1990.

[12] A. Marzo, and B. W. Drinkwater, "Holographic Acoustic Tweezers", *Proceedings of the National Academy of Sciences*, vol. 116, no. 1, pp. 84-89, 2019.

[13] M. Wu, A. Ozcelik, J. Rufo, Z. Wang, R. Fang, and T. J. Huang, "Acoustofluidic separation of cells and particles", *Microsystems & nanoengineering*, vol. 5, no. 1, pp. 1-18, 2019.

[14] M. Wiklund, R. Green, and M. Ohlin, "Acoustofluidics 14: applications of acoustic streaming in microfluidic devices", *Lab on a Chip*, vol. 12, no. 14, pp. 2438-2451, 2012.

[15] A. A. Doinikov, "Acoustic radiation forces: classical theory and recent advances", *Recent research developments in acoustics*. Vol. 1, pp. 39–67, 2003.

[16] A. Wixforth, C. Strobl, C. Gauer, A. Toegl, and J. Scriba, "Acoustic manipulation of small droplets", *Analytical and bioanalytical chemistry*, vol. 379, no. 7, pp. 982–991, 2004.

[17] L. Schmid, D. A. Weitz, and T. Franke, "Sorting drops and cells with acoustics: acoustic microfluidic fluorescence-activated cell sorter", *Lab on a Chip*, vol. 14, no. 19, pp. 3710–3718, 2014.

[18] K. Melde, A. G. Mark, T. Qiu, and P. Fischer, "Holograms for acoustics", *Nature*, vol. 537, no. 7621, pp. 518-522, 2016.

[19] M. A. Andrade, G. D. Skotis, S. Ritchie, D. R. Cumming, M. O. Riehle, and A. L. Bernassau, "Contactless Acoustic Manipulation and Sorting of Particles by Dynamic Acoustic Fields", *IEEE Transactions on Ultrasonics, Ferroelectrics, and Frequency Control*, vol. 63, no.10, pp. 1593-1600, 2016.

[20] Y. Yang, T. Ma, S. Li, Q. Zhang, J. Huang, Y. Liu, and H. Zheng, "Self-Navigated 3D Acoustic Tweezers in Complex Media Based on Time Reversal", *Research*, pp. 1-9, 2021.

[21] Q. Hu, T. Ma, Q. Zhang, J. Wang, Y. Yang, F. Cai, and H. Zheng, "3-D Acoustic Tweezers Using a 2-D Matrix Array with Time-Multiplexed Traps", *IEEE Transactions on Ultrasonics, Ferroelectrics, and Frequency Control*, vol. 68, no. 12, pp. 3646-3653, 2021.

[22] Y. Tang, S. Liu, and E. S. Kim, "MEMS focused ultrasonics transducer with air-cavity lens based on polydimethylsiloxane (PDMS) membrane", *the 33$^{rd}$ IEEE International Conference on Micro Electromechanical Systems*, Vancouver, Canada, 2020, pp. 58-61.

[23] M. A. Ghanem, A. D. Maxwell, Y. N. Wang, B. W. Cunitz, V. A. Khokhlova, O. A. Sapozhnikov and M.R. Bailey, "Noninvasive acoustic manipulation of objects in a living body", *Proceedings of the National Academy of Sciences (PNAS)*, vol. 117, no. 29, pp, 16848-16855, Jul. 2020.

[24] C. Zhong, Y. Jia, D. C. Jeong, Y. Guo, and S. Liu, "AcousNet: A Deep Learning based Approach to Dynamic 3D Holographic Acoustic Field Generation from Phased Transducer Array", *IEEE Robotics and Automation Letters*, vol. 7, no. 2, pp. 666-673, 2021.

[25] S. Liu, Y. Jia, Y. F. Li, Y. Guo, and H. Lu, "Simultaneous Precision Assembly of Multiple Objects through Coordinated Micro-robot Manipulation", *IEEE International Conference on Robotics and Automation*, pp. 6280-6286, 2021.

[26] S. Liu, D. Xu, D. Zhang, and Z. Zhang, "High Precision Automatic Assembly Based on Microscopic Vision and Force Information", *IEEE Transactions on Automation Science and Engineering*, vol. 13, no. 1, pp. 382–393, 2016.